# ALLNet: A Hybrid Convolutional Neural Network to Improve Diagnosis of Acute Lymphocytic Leukemia (ALL) in White Blood Cells

Sai Mattapalli*

Thomas Jefferson High School
for Science & Technology
saimattapalli1@gmail.com
Orcid: 0000-0001-7096-0208

Rishi Athavale*

Academy of Engineering
and Technology
rishi.athavale1@gmail.com
Orcid: 0000-0001-5271-1937

*Denotes equal contribution to writing paper regardless of the order of names

***Abstract*- Due to morphological similarity at the microscopic level, making an accurate and time-sensitive distinction between blood cells affected by Acute Lymphocytic Leukemia (ALL) and their healthy counterparts calls for the usage of machine learning architectures. However, three of the most common models, VGG (see Fig. 3), ResNet (see Fig. 4), and Inception (see Fig. 5), each come with their own set of flaws with room for improvement which demands the need for a superior model. ALLNet, the proposed hybrid convolutional neural network architecture, consists of a combination of the VGG, ResNet, and Inception models. The ALL Challenge dataset of ISBI 2019 (available here) contains 10,691 images of white blood cells which were used to train and test the models. 7,272 of the images in the dataset are of cells with ALL and 3,419 of them are of healthy cells. Of the images, 60% were used to train the model, 20% were used for the cross-validation set, and 20% were used for the test set. ALLNet outperformed the VGG, ResNet, and the Inception models across the board, achieving an accuracy of 92.6567%, a sensitivity of 95.5304%, a specificity of 85.9155%, an AUC score of 0.966347, and an F1 score of 0.94803 in the cross-validation set. In the test set, ALLNet achieved an accuracy of 92.0991%, a sensitivity of 96.5446%, a specificity of 82.8035%, an AUC score of 0.959972, and an F1 score of 0.942963. The utilization of ALLNet in the clinical workspace can better treat the thousands of people suffering from ALL across the world, many of whom are children [6].**

## 1. Introduction

Leukemias are a subset of cancer that stem from the excessive proliferation of would-be blood cells, usually white blood cells. A mutation that either activates oncogenes or turns off tumor suppressor genes, leads to uncontrolled iterations of the cell cycle and invariably cell division before cells are fully matured [12]. As the cells are divided before being fully developed, healthy white blood cells are replaced with underdeveloped ones which result in the body's defense against infections and diseases weakening. Acute Lymphocytic Leukemia, or ALL, is a type of leukemia that is acute, or fast-growing. This is why it proves to be fatal even after only a few months, which makes quick and accurate diagnosis imperative). Additionally, ALL is initiated in lymphoid cells instead of myeloid cells like some other forms of leukemia as the name suggests. ALL often starts in the bone marrow where new blood cells are formulated but can spread throughout the body, even affecting the liver and central nervous system.

Children five years old and younger are at the greatest risk of contracting ALL, the most common type of cancer in children, which further highlights how imperative it is that ALL gets the attention it deserves in our fight against pediatric cancers [5]. Symptoms of ALL can include weight loss, a loss of one's appetite, recurring infections, pale skin, and more [1].

There are numerous different ways that a person can be tested for ALL. Since ALL begins in the bone marrow, doctors have multiple different tests and methods for identifying ALL in an individual's bone marrow, such as bone marrow biopsy and bone marrow aspiration [1]. Both of these tests may result in the patient briefly feeling pain [1]. As previously mentioned, the quality and level of healthy blood cell production in the body can also be affected by ALL [1], and so blood tests are another way to test for ALL [1]. A doctor may also choose to administer a bone marrow test [1]. Additionally, there are also a variety of imaging methods that doctors can use, such as x-rays, MRI scans, CT scans, and ultrasound [1]. Imaging tests have an advantage over other types of tests due to their lack of radiation and their ease of administration [1]. Due to a large number of ALL cases a year and its fast rate of progression throughout the body [1], a simple method to be able to test for and identify ALL as quickly as possible is essential to fighting it. This study aims to create ALLNet, an automated method for identifying if a given image of a cell is an ALL cell so that, once the images of a patient's blood cells are acquired, a doctor can briskly obtain a diagnosis.

## 2. Methods

*2.1 Dataset*

For this study, the machine learning models were trained and evaluated by using data from the ALL Challenge dataset of ISBI 2019. Each image was labeled by an expert based on their domain knowledge. The dataset includes data from 118 participants and 118 studies and includes a total of 10,691 images in the training set, with 7,272 of those images being of an ALL cell and 3,419 of those images being of a healthy cell. Each image is stored as a 450x450x3 array. Each label is either 0 or 1, with a label of 0 denoting that it is a healthy cell and a label of 1 denoting that it is an ALL cell. Example images are shown in Fig. 1.A and Fig. 1.B.

*2.2 Data Preprocessing*

For the dataset, images from the original dataset's training set were taken and split into a training, cross-validation, and test set. They were split into a training set using 60% of the images (6414 images), a cross-validation set using 20% of the images (2138 images), and a test set using 20% of the images (2139 images). The data was then standardized by subtracting the mean of the training set images from all of the images in the dataset and then dividing all of the images in the dataset by the standard deviation of the training set images. This results in the training set having a mean of zero and a unit standard deviation. Standardization decreases the values of the inputs while preserving information, which allows the model to more easily fit the data. The mean and standard deviation were taken from the training set rather than the entire dataset to prevent data leakage, in which data from the cross-validation and test set appear in the training set. This can result in the model fitting to data from the cross-validation and test set, which would make the cross-validation and test set performance a poor metric of the model's performance on new data. The images are also resized to 250x250x3 arrays. Due to a large number of images, it would be inefficient to have them all loaded into memory. Instead, an ImageGenerator class was created which stores the file names and labels for a dataset and then only loads the number of images needed for a single batch. When these images are loaded by the ImageGenerator, they are also standardized and resized. Example images are shown in Fig. 2.A and Fig. 2.B.

A B

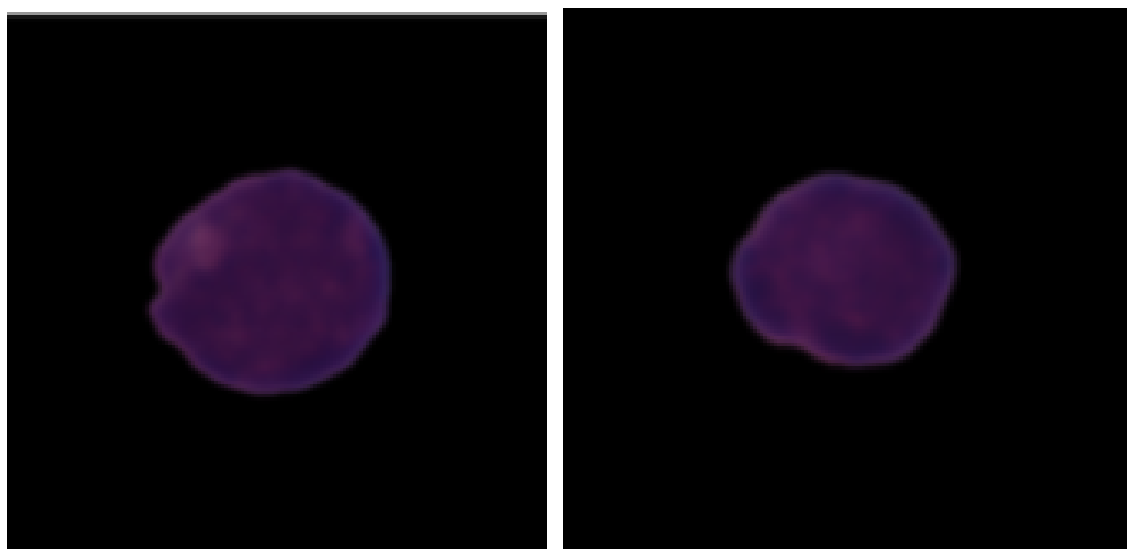

Fig 1.A. Image of a blood cell with ALL from the dataset.
Fig 1.B. Image of a healthy blood cell from the dataset.

A B

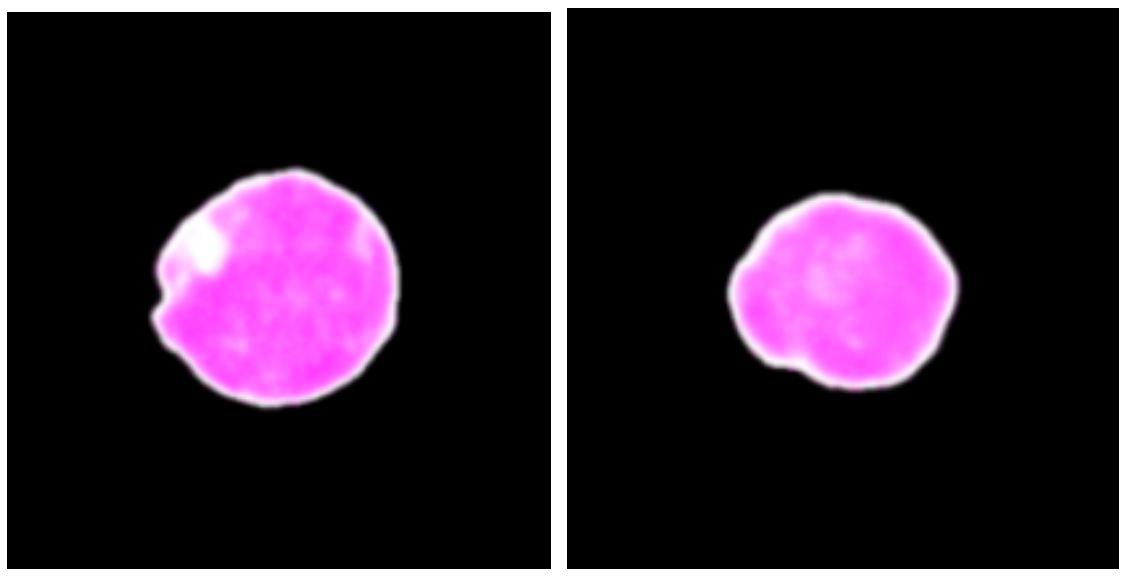

Fig 2.A. Preprocessed image of a blood cell with ALL from the dataset (standardized and resized to a 250x250 image).
Fig 2.B. Preprocessed image of a healthy blood cell from the dataset (standardized and resized to a 250x250 image).

*2.3 Model Architecture*

Three common convolutional neural networks (CNNs) for ALL classification and general computer vision tasks are VGG16, ResNet50, and InceptionV3. Their pre-trained weights from the ImageNet database were used to create a VGG, ResNet, and Inception model, which were trained on the training set using 40 epochs and a batch size of 32. Using pre-trained weights is known as transfer learning and allows for a model to transfer knowledge learned from a previous dataset to a new problem. The models were trained to minimize their binary cross-entropy loss on the training set, which is given by the following equation:

$$J(\theta) = -\frac{1}{m}\left[\sum_{i=1}^{m} y^{(i)} \log \hat{y}^{(i)} + \left(1 - y^{(i)}\right)\left(1 - \hat{y}^{(i)}\right)\right] \quad (1)$$

Equation (1) gives the binary cross-entropy loss where J($\theta$) is the binary cross-entropy loss, $\theta$ is the parameters of the model, $y^{(i)}$ is the ith label in the training set, $x^{(i)}$ is the ith input in the training set, and $\hat{y}^{(i)}$ is the model's prediction for the ith input in the training set.

The models were also trained using checkpoints, which saved the model at the epoch in which it had the highest accuracy on the cross-validation set. This procedure was implemented using Python and TensorFlow/Keras in Google Colab using 27.3 GB of RAM and a GPU.

*2.4 VGG*

VGG16 is a CNN architecture that achieved a top-1 accuracy of 0.713 and a top-5 accuracy of 0.901 on the ImageNet database. It is a relatively simple architecture and includes a series of 3x3 convolutions and max-pooling layers. The architecture for the model is shown in Fig. 3.

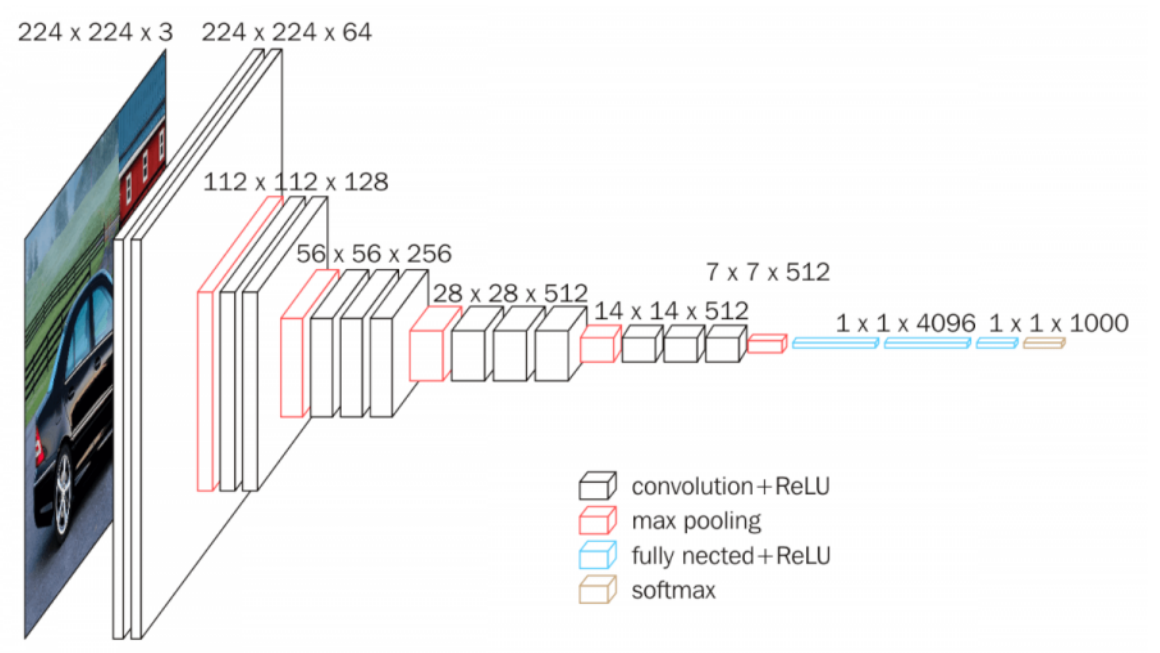

Fig 3. VGG16 architecture [13].

The VGG model used was built on a VGG16 model pre-trained on the ImageNet database. The top of the VGG16 model was removed and two convolutional layers that each had 512 1x1 kernels, a 2x2 max-pooling layer, and five fully connected layers were added. The output layer has a single neuron and uses the sigmoid activation function. During training, the first ten layers of the network were not trained.

*2.5 ResNet*

ResNet50 is a CNN architecture that achieved 0.749 top-1 accuracies and 0.921 top-5 accuracies on the ImageNet database. It utilizes a series of residual blocks, in which activations from previous layers are added to the activations of future layers. This helps prevent vanishing gradients (in which gradients get extremely close to zero towards the earlier layers of the network), which allows for deeper networks to be trained without losing performance. The architecture for the model is shown in Fig 4.

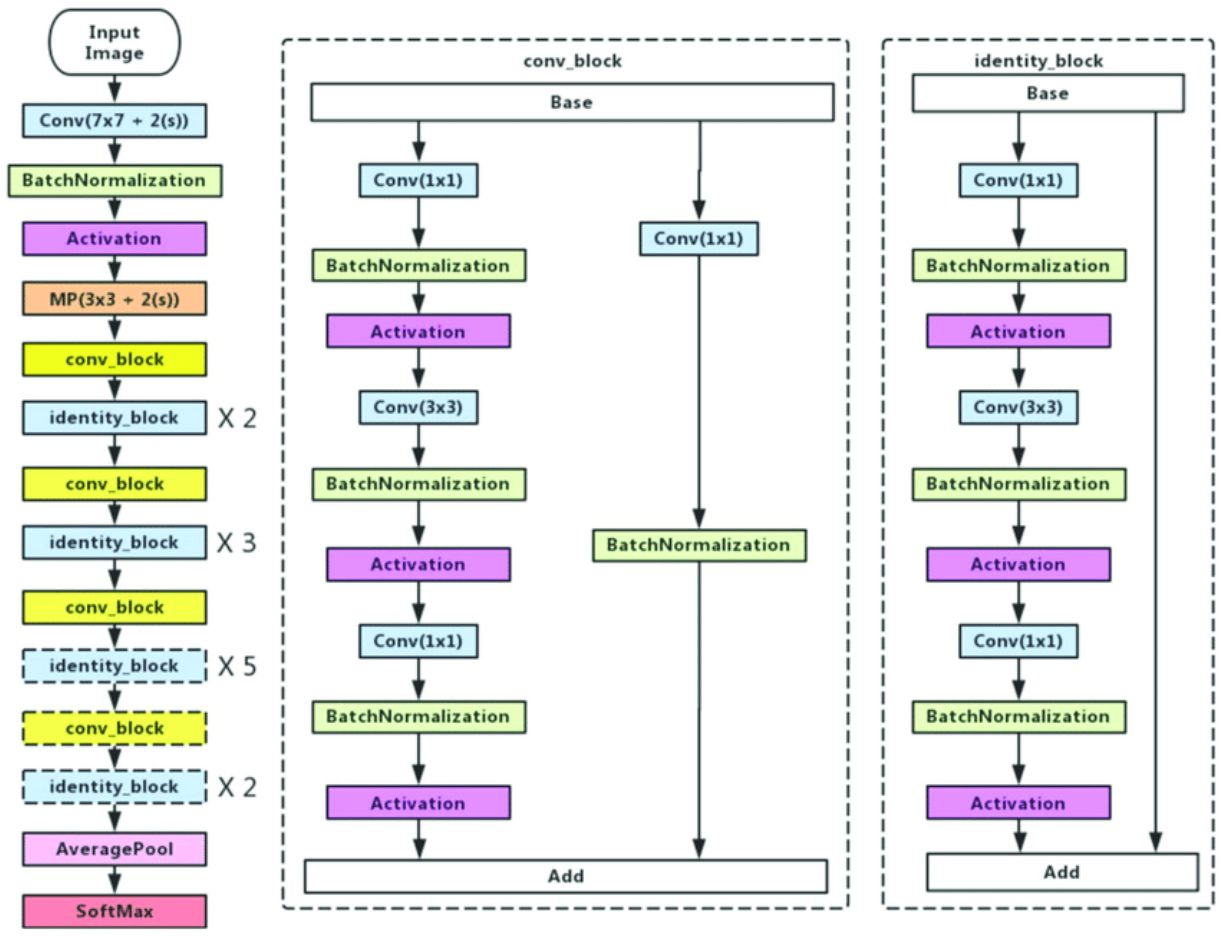

Fig 4. ResNet50 architecture [4].

*2.6 Inception*

InceptionV3 is a CNN architecture that achieved 0.779 top-1 accuracy and 0.937 top-5

accuracies on the ImageNet database. The InceptionV3 model applies multiple convolutional layers with different sized kernels and then concatenates them together. This lets the network try multiple different filter sizes rather than the programmer having to handpick them. The architecture for the model is shown in Fig 5.

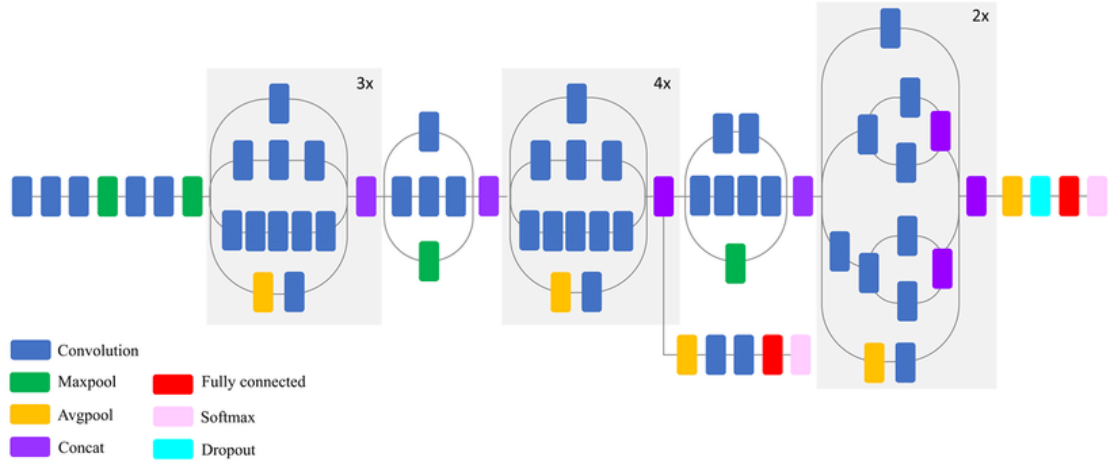

Fig 5. InceptionV3 architecture [8].

### *2.7 Proposed Model*

ALLNet is a hybrid CNN that combines InceptionV3, VGG16, and ResNet50 models that have been pre-trained on the ImageNet database. All of these model architectures are extremely powerful on their own, especially since they've already been pre-trained on the ImageNet database. However, this task is arduous because the ALL cells look almost identical to the healthy cells, and it is even more difficult in this case due to the relatively small number of images available to train the models on as a result of limitations on memory. To make a model with the ability to understand the subtle differences between the ALL cells and the healthy cells, a hybrid model that combines the strengths of each model is proposed so that the hybrid model will be able to gain a broader understanding of the data and the differences between the ALL cells and the healthy cells. The tops of those models were removed and their activations are concatenated into 'concatenate2'. The activations of two layers from each of the combined models are passed through a max-pooling layer, a series of convolutional layers, and then concatenated into 'concatenate1'. The 'concatenate1' layer was then passed into two 1x1 convolutional layers. It is then concatenated with 'concatenate2' into 'concatenate3'. This layer is then passed through a 3x3 max pooling layer with a stride of 2, a 1x1 convolutional layer, and then two fully connected layers. The output layer has a single neuron and uses the sigmoid activation function.

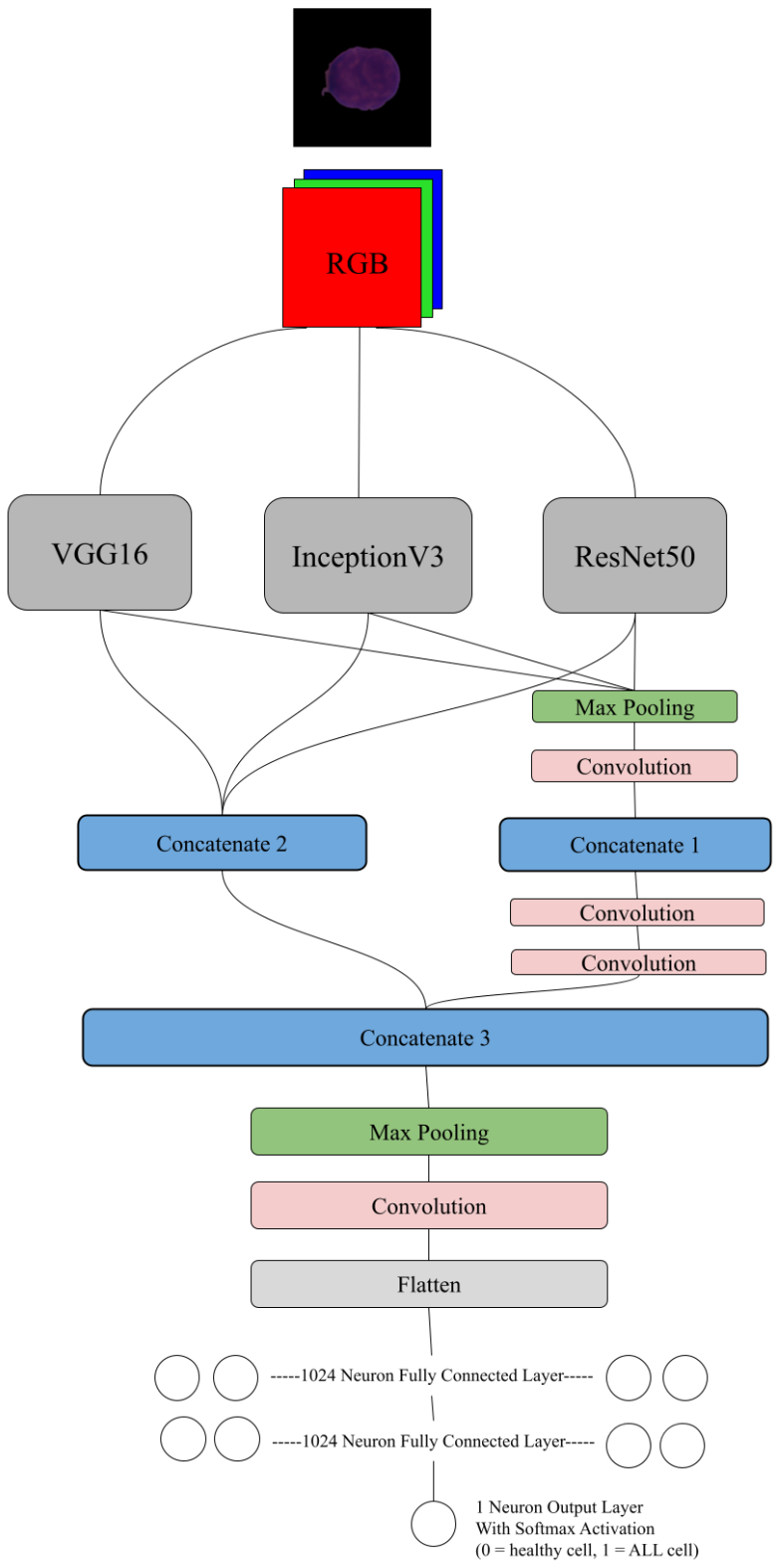

Fig. 6. Architecture of the proposed model ALLNet

This model was trained using the same settings as the other models

## 3. Results

After training, the models were evaluated on the cross-validation set using four metrics: accuracy, sensitivity, specificity, AUC score, and F1 score. The results for each model on these metrics are shown in Table 1.

TABLE I
ACCURACY, SENSITIVITY, SPECIFICITY, AUC SCORE, AND F1 SCORE FOR EVERY MODEL ON THE CROSS-VALIDATION SET

| Model | Accuracy (%) | Sensitivity (%) | Specificity (%) | AUC | F1 |
|---|---|---|---|---|---|
| VGG | 86.2021 | 91.2608 | 74.3349 | 0.906 | 0.90267 |
| ResNet | 86.015 | 92.0614 | 71.8310 | 0.899 | 0.90225 |
| Inception | 70.1123 | **100** | 0 | 0.5 | 0.82430 |
| ALLNet | **92.6567** | 95.5304 | **85.9155** | **0.966** | **0.94803** |

As shown in Table I, the proposed model ALLNet generally outperformed all of the other models, as it had the highest accuracy, the second-highest sensitivity, the highest specificity, the highest AUC score, and the highest F1 score. Now that ALLNet has been determined to have performed the best on the cross-validation set, it will now be tested on the test set. The results for ALLNet on the test set are shown in Table II.

TABLE II
ACCURACY, SENSITIVITY, SPECIFICITY, AUC SCORE, AND F1 SCORE FOR ALLNET ON THE TEST SET

| Model | Accuracy (%) | Sensitivity (%) | Specificity (%) | AUC | F1 |
|---|---|---|---|---|---|
| ALLNet | 92.0991 | 96.5446 | 82.8035 | 0.960 | 0.94296 |

## 4. DISCUSSION & CONCLUSION

Acute lymphocytic leukemia is one of the most common cancers among children and about 3,000 new cases of it arise every year in the United States alone [5]. One of the things that makes it so deadly is its rapid rate of spread, and it can be deadly in mere months if not properly treated [16]. Like many other cancers, but especially for ALL, having it be diagnosed early on is significant to fighting it. While there are many ways to diagnose it, such as bone marrow tests, blood tests, and chromosome tests, arguably the easiest and most efficient way is through imaging. To make it quicker and more effective to diagnose ALL through imaging, especially for countries with a small number of doctors relative to their population, this study aims to develop ALLNet, a hybrid convolutional neural network model that combines the VGG16, ResNet50, and InceptionV3 models. After training on 6,414 images from the ALL Challenge dataset of ISBI 2019, ALLNet will be able to classify a cell as either being an ALL cell or being a healthy cell better than the VGG, ResNet, and Inception models with high accuracy. On the cross-validation set, which had 2138 images, ALLNet achieved an accuracy of 92.6567%, a sensitivity of 95.5304%, a specificity of 85.9155%, an AUC score of 0.966347, and an F1 score of 0.94803, which surpassed the performance of the other models. The model was then able to achieve an accuracy of 92.0991%, a sensitivity of 96.5446%, a specificity of 82.8035%, an AUC score of 0.959972, and an F1 score of 0.942963 on the test set, which had 2139 images, making it suitable for ALL classification in hospitals.

Other methods that have the potential to improve on ALLNet's current performance include increasing the overall size of the dataset, training the models on more epochs, tuning the number of layers that are allowed to be trained in the pre-trained models, and using models pre-trained on different datasets. Increasing the size of the dataset would help prevent the model from overfitting, which could result in better cross-validation and test set performance. The dataset could be increased by gathering more images, however, it could take a large amount of time to get enough images for significant improvement. A better approach to increasing dataset size would likely be through data augmentation, in which data is altered to artificially create new data. While this may not add as much diverse information to the dataset as new images would, it would be a simple and quick approach to increasing the size of the dataset. Increasing the number of epochs could also result in better performance, as our procedure trained the models on only 40 epochs. However, this approach may not lead to significantly better results on the cross-validation set or the test set because the cross-validation loss and accuracy of the models generally didn't seem to improve very much with greater epochs. In this procedure, several layers in each model were prevented from being trained. Tuning whichever layers are allowed to be trained could help the models learn more useful features, which would improve performance. There are also a variety of other datasets that models could be pre-trained on that could result in better performance in classifying ALL cells. In particular, medical imaging datasets could be useful for pretraining as the features that models could learn from those datasets are more likely to be relevant to classifying ALL cells.

Based on its performance on a relatively small training set of 6414 images, it is very possible that ALLNet, especially after undergoing some slight modifications, has the potential to greatly improve the efficiency and speed of diagnosing ALL across the world.